\documentclass{article}
\usepackage{spconf,amsmath,graphicx,booktabs}
\usepackage[hidelinks]{hyperref}

\title{Dataset-Dependent Effects of Cross-Depth Aggregation and Soft-Routed Experts in EEG Foundation Model Fine-Tuning}
\name{Mingyang Jiang$^{1}$, Yamin Li$^{1}$, Daniel Moyer$^{1}$, Fan Ma$^{2}$, Hua Xu$^{2}$, and Catie Chang$^{1}$}
\address{$^{1}$Department of Computer Science, Vanderbilt University\\
$^{2}$Department of Biomedical Informatics and Data Science, Yale University}

\begin{document}
\maketitle

\begingroup
\renewcommand{\thefootnote}{}
\footnotetext{This work has been submitted to the IEEE for possible publication.
Copyright may be transferred without notice, after which this version may no longer be accessible.}
\endgroup

\begin{abstract}
EEG decoding tasks can rely on different temporal dynamics and cross-channel
relationships. We test whether specialized modules improve a fully fine-tuned
EEG foundation model by augmenting CBraMod with cross-depth Attention Residuals
(AttnRes) and two soft-routed expert banks. Across matched three-seed
experiments on FACED, ISRUC, SEED-V, and PhysioNet-MI, the complete model changes
mean balanced accuracy relative to full fine-tuning by $-0.12$, $+1.27$,
$+0.77$, and $-1.27$ points, respectively. AttnRes alone improves mean
balanced accuracy on three datasets, whereas adding experts on top of AttnRes
helps only FACED and SEED-V. These gains come with substantial overhead: AttnRes
requires 2.11--2.88$\times$ runtime and 1.78--2.67$\times$ memory, while the
complete model requires 2.41--3.04$\times$ runtime and 1.86--2.85$\times$
memory. Overall, the added modules produce dataset-dependent, sometimes
opposing effects rather than consistent gains over full fine-tuning.
\end{abstract}


\section{Introduction}
\label{sec:intro}

Electroencephalography (EEG) records multichannel neural activity with high
temporal resolution and supports diverse decoding tasks, including emotion
recognition, sleep staging, and motor imagery. Decoding these signals requires
modeling patterns distributed across time and channels, but the relevant
structure can differ across tasks and datasets. Conventional EEG models
therefore often include explicit temporal and spatial processing
\cite{lawhern2018eegnet,song2023eegconformer}.

Recent EEG foundation models pretrain large encoders on broad EEG corpora and
transfer their representations across downstream tasks. Models such as CBraMod,
LaBraM, and BrainGPT demonstrate this paradigm for transferable EEG
representation learning \cite{wang2025cbramod,jiang2024labram,yue2024braingpt}.
When adapting such a pretrained encoder, full fine-tuning allows all backbone
parameters to change. This raises a practical question: do additional
specialized modules still help, or can a fully trainable backbone learn similar
transformations on its own?

We investigate this question with CBraMod using two complementary mechanisms
motivated by the variability of EEG decoding. First, cross-depth Attention
Residuals (AttnRes) combine representations from earlier encoder layers,
allowing downstream processing to draw on features learned at different depths
rather than relying only on the standard residual path. This may be useful when
different EEG tasks depend on different temporal or cross-channel patterns.
Second, soft-routed expert banks use input-dependent routing weights to combine
several nonlinear expert networks, allowing the model to apply different
transformations to different EEG inputs. Related expert-routing methods have
recently been explored for EEG representation learning \cite{ma2026trace}. We
therefore compare full fine-tuning, full fine-tuning with AttnRes, and the
complete model with AttnRes and experts to test whether these mechanisms provide
complementary benefit beyond a fully trainable backbone.

Our contributions are threefold. First, we test whether these additional
modules provide further benefit when a pretrained EEG backbone is already fully
trainable. Second, we separate the effects of cross-depth aggregation and
soft-routed expert capacity using matched three-seed comparisons across emotion
recognition, sleep staging, and motor imagery. Third, we characterize
dataset-dependent performance changes together with computational cost and
class- and subject-level effects.

\begin{figure*}[!t]
\centering
\includegraphics[width=0.84\textwidth]{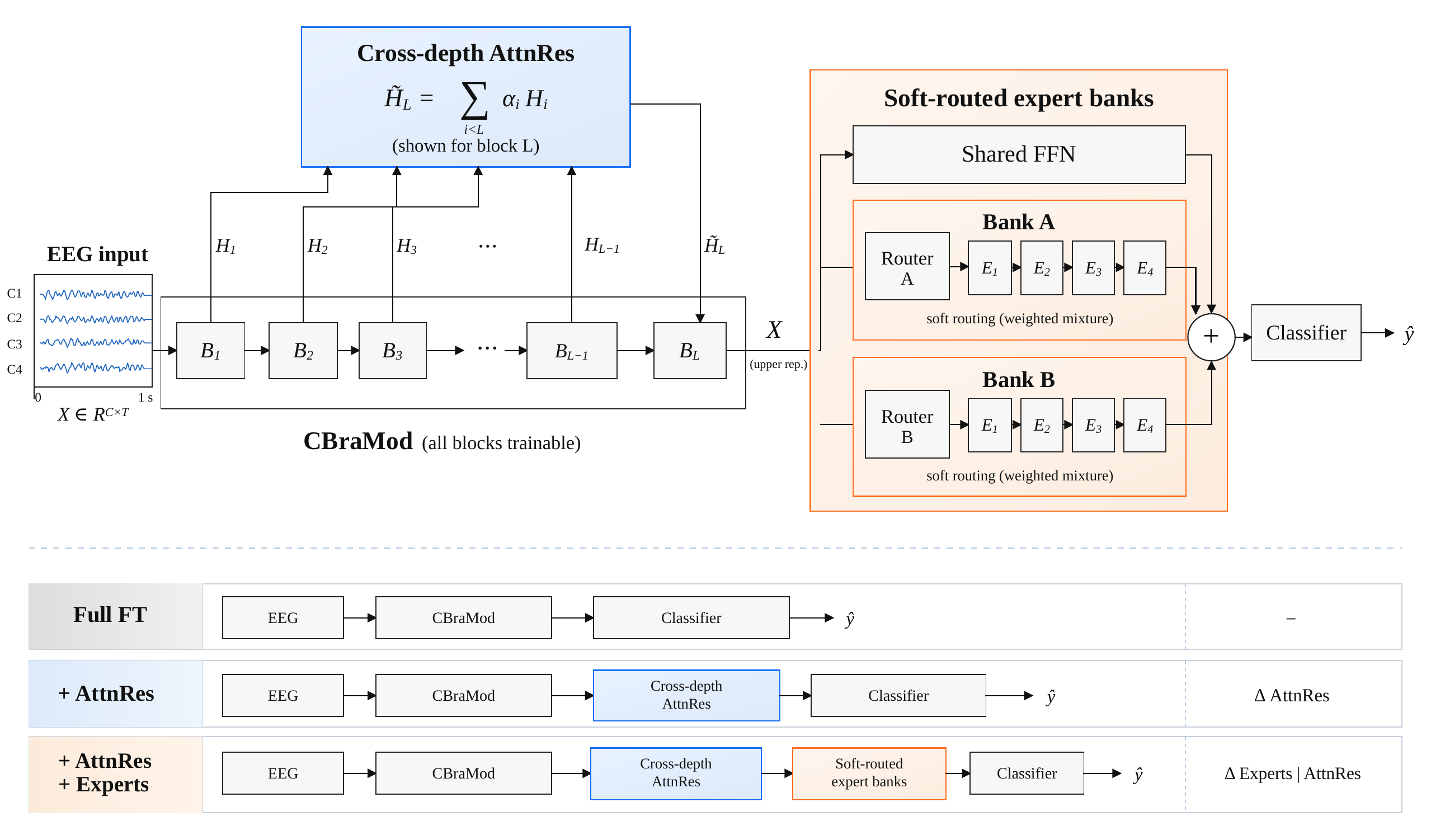}
\vspace{-12pt}
\caption{Cross-depth expert augmentation and staged comparisons. Cross-depth
AttnRes combines earlier CBraMod states before the final representation is
processed by two separate soft-routed expert banks in parallel with the shared
FFN. The lower panel summarizes the evaluated stages: Full FT, +AttnRes, and
+AttnRes + Experts.}
\vspace{-12pt}
\label{fig:method}
\end{figure*}

\section{Cross-Depth and Soft-Routed Expert Capacity}
\label{sec:method}

Our baseline fully fine-tunes CBraMod. In the augmented models, the backbone
remains trainable while we add cross-depth Attention Residuals (AttnRes) and two
separate expert banks. AttnRes learns a weighted combination of earlier block
outputs instead of using only the standard residual path
\cite{chen2026attentionresiduals}.

\subsection{Cross-depth aggregation}

CBraMod processes the EEG representation through a stack of encoder blocks.
For target block $i$, let $H_j\in\mathbf{R}^{B\times C\times S\times D}$
denote an earlier representation that can be used as a source, and let
$\mathcal{S}_i$ denote the available sources, including the patch embedding and
outputs of preceding blocks. We form
\begin{equation}
\begin{aligned}
\alpha_{i,j} &= \operatorname{softmax}_{j\in\mathcal{S}_i}\!\left(q_i^{\top}
\operatorname{RMSNorm}_i(H_j)\right),\\[-2pt]
\widetilde{H}_i &= \sum_{j\in\mathcal{S}_i} \alpha_{i,j} H_j,
\end{aligned}
\label{eq:attnres}
\end{equation}
Here $q_i\in\mathbf{R}^{D}$ and $\operatorname{RMSNorm}_i$ are learned per
target block (400 parameters per block, 4,800 across 12 blocks). In the
ungated setting, $\widetilde{H}_i$ replaces the ordinary input before
$\operatorname{norm1}$ and self-attention, which then applies its usual
residual update. The softmax weights are computed across the available depth
states at each position, and aggregation begins at encoder block 0 rather than
only at final pooling.

\subsection{Soft-routed expert capacity}

At the top (final) encoder layer, let $B$ denote the baseline pre-FFN representation
and $A$ its AttnRes-augmented counterpart. We form one routing feature vector per
sample, $u=[\mu(B);\mu(A);\mu(A-B)]$, where $\mu$ averages the channel and patch
axes. The implemented router uses all three terms, including the difference
$A-B$.
Let $X$ be the layer-normalized
pre-FFN representation supplied to the
experts. Each expert is a 200--800--200 GELU MLP using the dropout rate specified
by the dataset protocol.
For each bank $b\in\{A,B\}$, the model computes
{\small
\begin{equation}
\begin{aligned}
p^{(b)}&=\operatorname{softmax}(r_b(u)/\tau),\qquad
R_b(X)=\sum_{e=1}^{4}p^{(b)}_eE^{(b)}_e(X),\\
F(X)&=F_{\rm shared}(X)+R_A(X)+R_B(X),
\end{aligned}
\label{eq:experts}
\end{equation}
}
with $\tau=1.5$. Here $F_{\rm shared}$ is the ordinary dense FFN output. The
two banks use separate routers and expert parameters but receive the same
normalized token representation. They are separate learned parameter banks,
not fixed spatial or spectral operators. Each routing vector
is computed once per sample by averaging channel and patch axes and is shared
across that sample's tokens. The router forms a weighted mixture of all four
experts in each bank. We use dense routing, so every expert contributes to each
sample according to its routing weight. This avoids additional effects from
sparse expert dispatch or load-balancing mechanisms
\cite{shazeer2017moe,fedus2022switch}.

\section{Experimental Protocol}
\label{sec:protocol}

We evaluate FACED and SEED-V for emotion recognition, ISRUC for sleep staging,
and PhysioNet-MI for motor imagery
\cite{chen2023faced,liu2021seedv,khalighi2016isruc,schalk2009eegmmidb,goldberger2000physiobank}.
Table~\ref{tab:protocol} summarizes the fixed dataset splits and training
schedules. Within each pair, both conditions use the same pretrained CBraMod
checkpoint, preprocessing, split, classifier, and checkpoint-selection rule.
For ISRUC, we use the same ordered split of subjects 1--80/81--90/91--100, and it
is unchanged across paired conditions. SEED-V retains the fixed CBraMod LMDB
benchmark protocol for comparability and is not subject-disjoint.

\begin{table*}[t]
\centering
\small
\setlength{\tabcolsep}{2.5pt}
\caption{Dataset and training protocols used in the matched comparisons.
Ch., EEG channels; BS, batch size. $^{*}$SEED-V uses the fixed CBraMod LMDB
benchmark split rather than a subject-disjoint partition.}
\label{tab:protocol}
\begin{tabular}{l l c c l c l}
\toprule
Dataset & Task & Ch. & Classes & Split protocol &
Subjects (Tr/Val/Te) & Epoch / BS / LR \\
\midrule
FACED & Emotion & 32 & 9 & Fixed subject/session &
86 / 18 / 19 & 40 / 32 / $2{\times}10^{-4}$ \\
ISRUC & Sleep staging & 6 & 5 & Ordered subject split &
80 / 10 / 10 & 30 / 16 / $3{\times}10^{-5}$ \\
SEED-V & Emotion & 62 & 5 & CBraMod LMDB benchmark &
--$^{*}$ & 25 / 64 / $3{\times}10^{-5}$ \\
PhysioNet-MI & Motor imagery & 64 & 4 & Subject-disjoint &
76 / 16 / 17 & 30 / 64 / $3{\times}10^{-5}$ \\
\bottomrule
\end{tabular}
\end{table*}

Each condition is run with three prespecified random seeds, 3407, 2024, and
2027. The best checkpoint is selected using validation Cohen's $\kappa$ and
then evaluated once on the held-out test split. We report balanced accuracy
(BA), weighted-F1 (wF1), and Cohen's $\kappa$. Scores in the result tables are
scaled by 100 for compact presentation, and all deltas are computed from
paired runs that use the same dataset protocol and seed. Because we use only
three prespecified seeds, we treat the paired differences as descriptive rather
than inferential.

\emph{Implementation details.} We use ungated pre-attention AttnRes from layer
0. Routers use a hidden width of 128, and expert output weights are
zero-initialized. All parameters follow the fixed dataset schedules in
Table~\ref{tab:protocol}; the classifier uses all patch representations.

End-to-end training wall-clock time and peak allocated CUDA memory are measured
in the same NVIDIA A6000 environment for matched runs. For diagnostics,
class-wise recall is averaged over matched seeds. For PhysioNet-MI, we
additionally analyze paired BA changes across the 17 held-out subjects and
estimate uncertainty by bootstrapping those subjects. Validation trajectories
use the checkpoint-selection criterion.

\section{Results and Discussion}
\label{sec:results}

Tables~\ref{tab:main_results} and~\ref{tab:components} summarize test
performance, paired augmentation effects, and computational overhead across
the four datasets.

\begin{figure*}[t]
\centering
\includegraphics[width=0.82\textwidth]{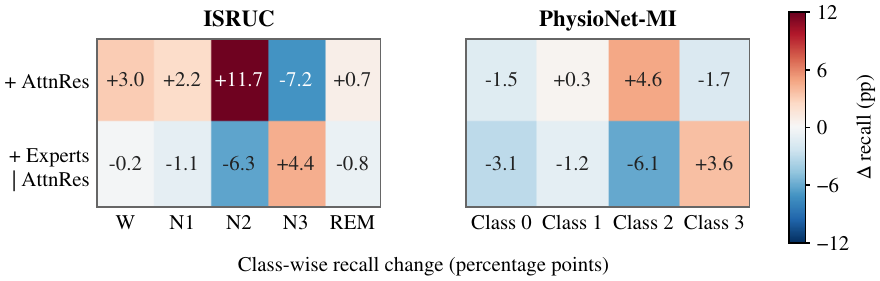}
\vspace{-4pt}
\caption{Mean changes in class-wise recall over three matched seeds. The first
row shows AttnRes minus Full FT and the second shows the complete model minus
AttnRes; values are descriptive percentage-point changes, and the shared
diverging color scale is centered at zero and spans $-12$ to $+12$.}
\label{fig:classwise}
\end{figure*}

\begin{table*}[t]
\centering
\small
\setlength{\tabcolsep}{4pt}
\caption{Test performance under matched full fine-tuning and cross-depth
expert augmentation. Each method cell reports BA / wF1 / $\kappa$ as mean
$\pm$ sample SD over seeds 3407, 2024, and 2027. All scores are multiplied by
100 for compact presentation; Full FT serves as the within-protocol reference
for each dataset.}
\label{tab:main_results}
\begin{tabular}{lp{0.36\textwidth}p{0.36\textwidth}}
\toprule
Dataset & Full FT (BA / wF1 / $\kappa$) & + AttnRes + Experts (BA / wF1 / $\kappa$) \\
\midrule
FACED & \textbf{58.56 $\pm$ 0.80 / 58.65 $\pm$ 0.84 / 53.03 $\pm$ 0.86} & 58.44 $\pm$ 0.44 / 58.48 $\pm$ 0.42 / 52.88 $\pm$ 0.47 \\
ISRUC & 77.52 $\pm$ 0.93 / 78.66 $\pm$ 1.25 / 72.56 $\pm$ 1.48 & \textbf{78.79 $\pm$ 1.02 / 80.56 $\pm$ 1.09 / 74.97 $\pm$ 1.31} \\
SEED-V & 39.36 $\pm$ 0.17 / 40.04 $\pm$ 0.18 / 24.74 $\pm$ 0.17 & \textbf{40.13 $\pm$ 0.67 / 40.97 $\pm$ 0.60 / 25.83 $\pm$ 0.90} \\
PhysioNet-MI & \textbf{63.11 $\pm$ 1.27 / 63.14 $\pm$ 1.25 / 50.80 $\pm$ 1.70} & 61.84 $\pm$ 1.02 / 61.86 $\pm$ 1.06 / 49.09 $\pm$ 1.35 \\
\bottomrule
\end{tabular}
\vspace{5pt}
\caption{Staged component comparison and cost. BA is mean $\pm$ SD over the
matched seeds and is scaled by 100. $\Delta$AttnRes denotes AttnRes $-$ Full
FT, and $\Delta$Experts$\mid$AttnRes denotes the complete model $-$ AttnRes.
The time and memory columns (C denotes the complete model) are ratios to Full
FT.}
\label{tab:components}
\begin{tabular}{lccccccc}
\toprule
Dataset & Full FT & + AttnRes & + AttnRes + Exp. & $\Delta_{\rm AttnRes}$ & $\Delta_{\rm Experts\mid AttnRes}$ & Time $\times$ (C) & Mem. $\times$ (C) \\
\midrule
FACED & 58.56 $\pm$ 0.80 & 56.97 $\pm$ 0.53 & 58.44 $\pm$ 0.44 & $-$1.58 $\pm$ 0.33 & +1.46 $\pm$ 0.95 & 2.41 & 1.86 \\
ISRUC & 77.52 $\pm$ 0.93 & 79.60 $\pm$ 0.39 & 78.79 $\pm$ 1.02 & \textbf{+2.08 $\pm$ 1.25} & $-$0.81 $\pm$ 0.77 & 2.91 & 2.85 \\
SEED-V & 39.36 $\pm$ 0.17 & 39.87 $\pm$ 0.40 & \textbf{40.13 $\pm$ 0.67} & +0.50 $\pm$ 0.25 & +0.26 $\pm$ 0.94 & 2.55 & 2.38 \\
PhysioNet-MI & 63.11 $\pm$ 1.27 & \textbf{63.54 $\pm$ 0.26} & 61.84 $\pm$ 1.02 & +0.43 $\pm$ 1.53 & $-$1.71 $\pm$ 1.27 & 3.04 & 2.49 \\
\bottomrule
\end{tabular}
\end{table*}

\subsection{Heterogeneous performance shifts}
\label{sec:task_results}

The complete model shows heterogeneous, dataset-dependent effects. ISRUC shows the largest positive
complete-model mean BA difference: all three paired seeds favor the complete
model, with mean BA higher by
$1.27\pm1.46$ points; weighted-F1
and $\kappa$ increase by $1.90\pm1.66$ and $2.40\pm1.89$ points. PhysioNet-MI
shows the opposite BA shift, $-1.27\pm0.26$ points, with all seeds negative.
FACED does not show a consistent advantage across the three matched seeds
($-0.12\pm1.24$ BA; one positive seed),
whereas SEED-V has a modestly higher mean ($+0.77\pm0.74$ BA; two positive seeds).
Across all four datasets, BA, weighted-F1, and $\kappa$ show the same direction
of change.

The best-performing stage therefore varies by dataset (Table~\ref{tab:components}).

\subsection{Component-wise incremental effects}
\label{sec:component_results}

Table~\ref{tab:components} defines $\Delta_{\rm AttnRes}$ as AttnRes minus
Full FT and $\Delta_{\rm Experts\mid AttnRes}$ as the complete model minus
AttnRes. AttnRes raises mean BA relative to Full FT on ISRUC, SEED-V, and
PhysioNet-MI, but lowers it on FACED. Adding experts on top of AttnRes decreases
mean BA on ISRUC and PhysioNet-MI but increases it on FACED and SEED-V;
weighted-F1 and $\kappa$ show the same directions. These opposing stage effects
show that comparing only Full FT with the final model can hide how AttnRes and
the expert module contribute differently. Because adding experts also increases
parameter count, $\Delta_{\rm Experts\mid AttnRes}$ measures the overall
incremental effect of the full expert module rather than isolating routing alone.

At the selected checkpoints, the entropy-based effective expert count, which
summarizes how many experts are meaningfully used, ranged from 2.57--4.00 across
datasets, seeds, and banks, while the dominant expert
received up to 0.91 of top-1 assignments. These inference-only router
statistics were not used for model selection.

\subsection{Class-wise stage effects}
\label{sec:classwise_results}
\vspace{-6pt}

To characterize which classes contribute to the aggregate changes, we re-evaluated
the selected test checkpoints and computed class-wise recall changes between
matched conditions.
Figure~\ref{fig:classwise} shows concentrated, dataset-specific class effects.
On ISRUC, AttnRes mainly increases N2 recall (+11.7 points) while reducing N3
($-7.2$), and the expert stage partially reverses both changes. On PhysioNet-MI,
experts reverse the AttnRes gain for class 2 and reduce recall for classes 0--1.
These patterns show that different classes gain and lose performance at
different stages rather than all classes changing uniformly.

On PhysioNet-MI, experts reduce BA for 12/17 held-out subjects (median paired
$\Delta$BA $=-1.98$; subject-bootstrap 95\% CI $[-3.29,-0.16]$). On ISRUC,
AttnRes remains above Full FT in validation $\kappa$ for all seeds across epochs
19--23 (+4.1 to +5.2 mean points), whereas the PhysioNet-MI expert disadvantage
is less consistent on validation.

\subsection{Accuracy--compute trade-off}
\label{sec:cost_results}

These accuracy changes incur substantial compute overhead. AttnRes adds only
4,800 parameters but incurs 2.11$\times$--2.88$\times$ runtime and
1.78$\times$--2.67$\times$ memory; the complete model adds 2.73M parameters and
incurs 2.41$\times$--3.04$\times$ runtime and 1.86$\times$--2.85$\times$ memory.
Because AttnRes retains intermediate activations, parameter count alone does not
capture its practical compute cost.

\section{Limitations and Conclusion}
\label{sec:conclusion}

This study is limited to one backbone, one protocol per dataset, and three
seeds; SEED-V is not subject-disjoint, and dataset differences conflate task and
split. The expert-stage comparison lacks a parameter-matched dense control. In
addition, ungated AttnRes changes the block input from the start of fine-tuning,
and the diagnostic analyses are descriptive.

Across the evaluated CBraMod settings, cross-depth aggregation and expert
capacity show dataset-dependent, sometimes opposing effects rather than consistent
gains over full fine-tuning. Thus, each added component should be evaluated
separately rather than only through a baseline-to-complete comparison.

\section*{Compliance with Ethical Standards}
This study is a secondary analysis of previously collected EEG data from
publicly available datasets. No new participants were recruited or contacted,
and no new participant-facing procedures were performed. The present analysis
used only previously collected, publicly available data and therefore did not
require additional ethical approval; ethical approval and informed-consent
procedures for the original data collections are documented in the corresponding
dataset publications and applicable access terms.

\section*{Funding and Conflicts of Interest}
We thank the Vanderbilt Institute for Surgery and Engineering (VISE) Summer
Research Program for funding support. The authors declare no conflicts of
interest.

\bibliographystyle{IEEEbib}
\bibliography{strings,refs}

@inproceedings{wang2025cbramod,
  author = {Wang, Jiquan and Zhao, Sha and Luo, Zhiling and Zhou, Yangxuan and Jiang, Haiteng and Li, Shijian and Li, Tao and Pan, Gang},
  title = {CBraMod: A Criss-Cross Brain Foundation Model for EEG Decoding},
  booktitle = {International Conference on Learning Representations},
  year = {2025}
}

@inproceedings{jiang2024labram,
  author = {Jiang, Wei-Bang and Zhao, Li-Ming and Lu, Bao-Liang},
  title = {Large Brain Model for Learning Generic Representations with Tremendous EEG Data in BCI},
  booktitle = {International Conference on Learning Representations},
  year = {2024}
}

@article{yue2024braingpt,
  author = {Yue, Tongtian and Gao, Xuange and Xue, Shuning and Tang, Yepeng and Guo, Longteng and Jiang, Jie and Liu, Jing},
  title = {BrainGPT: Unleashing the Potential of EEG Generalist Foundation Model by Autoregressive Pre-training},
  journal = {arXiv preprint arXiv:2410.19779},
  year = {2024}
}

@article{ma2026trace,
  author = {Ma, Fan and An, Qier and Chen, Peng and Qian, Lingfei and Lan, Xiang and Jiang, Mingyang and Gu, Zhiling and Papademetris, Xenophon and Xu, Hua},
  title = {TRACE: Temporal Routing with Autoregressive Cross-channel Experts for EEG Representation Learning},
  journal = {arXiv preprint arXiv:2605.11380},
  year = {2026}
}

@article{chen2026attentionresiduals,
  author = {{Kimi Team}},
  title = {Attention Residuals},
  journal = {arXiv preprint arXiv:2603.15031},
  year = {2026}
}

@article{lawhern2018eegnet,
  author = {Lawhern, Vernon J. and Solon, Amelia J. and Waytowich, Nicholas R. and Gordon, Stephen M. and Hung, Chou P. and Lance, Brent J.},
  title = {EEGNet: A Compact Convolutional Neural Network for EEG-Based Brain-Computer Interfaces},
  journal = {Journal of Neural Engineering},
  volume = {15},
  number = {5},
  pages = {056013},
  year = {2018},
  doi = {10.1088/1741-2552/aace8c}
}

@article{song2023eegconformer,
  author = {Song, Yonghao and Zheng, Qingqing and Liu, Bingchuan and Gao, Xiaorong},
  title = {EEG Conformer: Convolutional Transformer for EEG Decoding and Visualization},
  journal = {IEEE Transactions on Neural Systems and Rehabilitation Engineering},
  volume = {31},
  pages = {710--719},
  year = {2023},
  doi = {10.1109/TNSRE.2022.3230250}
}

@inproceedings{shazeer2017moe,
  author = {Shazeer, Noam and Mirhoseini, Azalia and Maziarz, Krzysztof and Davis, Andy and Le, Quoc V. and Hinton, Geoffrey and Dean, Jeff},
  title = {Outrageously Large Neural Networks: The Sparsely-Gated Mixture-of-Experts Layer},
  booktitle = {International Conference on Learning Representations},
  year = {2017}
}

@article{fedus2022switch,
  author = {Fedus, William and Zoph, Barret and Shazeer, Noam},
  title = {Switch Transformers: Scaling to Trillion Parameter Models with Simple and Efficient Sparsity},
  journal = {Journal of Machine Learning Research},
  volume = {23},
  number = {120},
  pages = {1--39},
  year = {2022}
}

@article{chen2023faced,
  author = {Chen, Jingjing and Wang, Xiaobin and Huang, Chen and Hu, Xin and Shen, Xinke and Zhang, Dan},
  title = {A Large Finer-Grained Affective Computing EEG Dataset},
  journal = {Scientific Data},
  volume = {10},
  number = {1},
  year = {2023},
  doi = {10.1038/s41597-023-02650-w}
}

@article{liu2021seedv,
  author = {Liu, Wei and Qiu, Jie-Lin and Zheng, Wei-Long and Lu, Bao-Liang},
  title = {Comparing Recognition Performance and Robustness of Multimodal Deep Learning Models for Multimodal Emotion Recognition},
  journal = {IEEE Transactions on Cognitive and Developmental Systems},
  volume = {14},
  number = {2},
  pages = {715--729},
  year = {2022},
  doi = {10.1109/TCDS.2021.3071170}
}

@article{khalighi2016isruc,
  author = {Khalighi, Sirvan and Sousa, Teresa and Santos, Jose Moutinho dos and Nunes, Urbano},
  title = {ISRUC-Sleep: A Comprehensive Public Dataset for Sleep Researchers},
  journal = {Computer Methods and Programs in Biomedicine},
  volume = {124},
  pages = {180--192},
  year = {2016},
  doi = {10.1016/j.cmpb.2015.10.013}
}

@article{schalk2009eegmmidb,
  author = {Schalk, Gerwin},
  title = {EEG Motor Movement/Imagery Dataset},
  journal = {PhysioNet},
  year = {2009},
  doi = {10.13026/C28G6P}
}

@article{goldberger2000physiobank,
  author = {Goldberger, Ary L. and Amaral, Luis A. N. and Glass, Leon and Hausdorff, Jeffrey M. and Ivanov, Plamen Ch. and Mark, Roger G. and Mietus, Joseph E. and Moody, George B. and Peng, Chung-Kang and Stanley, H. Eugene},
  title = {PhysioBank, PhysioToolkit, and PhysioNet: Components of a New Research Resource for Complex Physiologic Signals},
  journal = {Circulation},
  volume = {101},
  number = {23},
  pages = {e215--e220},
  year = {2000},
  doi = {10.1161/01.CIR.101.23.e215}
}

\end{document}